\PassOptionsToPackage{table}{xcolor}
\documentclass[10pt,twocolumn,letterpaper]{article}

\usepackage[pagenumbers]{cvpr} 

\definecolor{cvprblue}{rgb}{0.21,0.49,0.74}
\usepackage[pagebackref,breaklinks,colorlinks,allcolors=cvprblue]{hyperref}

\def\MethodName{DeepDeblurRF}
\def\SynthDataName{BlurRF-Synth}
\def\RealDataName{BlurRF-Real}
\def\SBDataName{BlurRF-SB}

\renewcommand{\paragraph}[1]{\vspace{2pt}\noindent\textbf{#1}~~}

\usepackage{kotex}
\usepackage[capitalize]{cleveref}
\usepackage{multirow}
\usepackage{multicol}
\usepackage{makecell}

\definecolor{skyblue}{rgb}{0.78, 0.88, 1.00}
\definecolor{red_blurry}{rgb}{0.95, 0.73, 0.75}
\definecolor{yellow_blurry}{rgb}{1.00, 1.00, 0.65}

\title{Exploiting Deblurring Networks for Radiance Fields}

\author{Haeyun Choi$^{1,\dag}$ \and Heemin Yang$^{2}$ \and Janghyeok Han$^{2}$ \and Sunghyun Cho$^{2}$ \\
\vspace{-4.5mm}
\and
$^1$KT \hspace{10mm} $^2$POSTECH \\
}

\begin{document}
\maketitle
{\let\thefootnote\relax\footnotetext{\noindent ${}^{\dag}$This work was done at POSTECH.}}
\begin{abstract}
    In this paper, we propose \MethodName{}, a novel radiance field deblurring approach that can synthesize high-quality novel views from blurred training views with significantly reduced training time. 
    \MethodName{} leverages deep neural network (DNN)-based deblurring modules to enjoy their deblurring performance and computational efficiency. 
    To effectively combine DNN-based deblurring and radiance field construction, we propose a novel radiance field (RF)-guided deblurring and an iterative framework that performs RF-guided deblurring and radiance field construction in an alternating manner.
    Moreover, \MethodName{} is compatible with various scene representations, such as voxel grids and 3D Gaussians, expanding its applicability. 
    We also present \SynthDataName{}, the first large-scale synthetic dataset for training radiance field deblurring frameworks.
    We conduct extensive experiments on both camera motion blur and defocus blur, demonstrating that \MethodName{} achieves state-of-the-art novel-view synthesis quality with significantly reduced training time.
\end{abstract}    
\section{Introduction}
\label{sec:intro}

\begin{figure*}[t!]
\centering
    \includegraphics[width=0.98\textwidth]{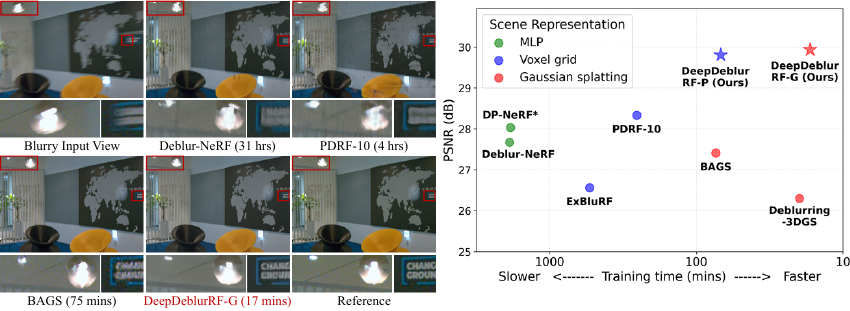}
    \caption{Given a set of multi-view blurry images even with non-linear outliers such as saturated pixels and noise, \MethodName{} performs high-quality novel-view synthesis with highly efficient training. \MethodName{}-P and \MethodName{}-G are the results of our framework, where radiance fields are constructed using Plenoxels~\cite{fridovich2022plenoxels} and 3D Gaussian Splatting~\cite{kerbl2023gaussiansplatting}, respectively. Note that DP-NeRF~\cite{lee2023dpnerf} was trained with two GPUs due to its memory demands, whereas other models were trained on a single NVIDIA TITAN RTX GPU.
    }
    \vspace{-3mm}
    \label{fig:teaser}
\end{figure*}

Novel-view synthesis has seen significant advancements in recent years, leading to impressive improvements in photo-realistic rendering quality. One key development in this field is the Neural Radiance Field (NeRF)~\cite{mildenhall2020nerf}, which leverages neural networks to generate highly detailed images from novel viewpoints. Building on this, various works have focused on reducing training time and enhancing rendering quality~\cite{mueller2022instantngp, chen2022tensorf, hu2023trimiprf, fridovich2022plenoxels}. More recently, 3D Gaussian Splatting (3DGS)~\cite{kerbl2023gaussiansplatting} has emerged, utilizing Gaussians as an explicit 3D representation. This technique enables high-quality scene reconstruction and real-time rendering through a differentiable rasterization method.

However, synthesizing sharp novel views from degraded training views remains a significant challenge. Capturing images in real-world conditions often leads to various degradations such as blur and noise.
Among them, blur makes it difficult to aggregate accurate 3D information from the training views, resulting in a blurry radiance field.
To address blur, several works~\cite{ma2022deblurnerf, wang2023badnerf, lee2023dpnerf, peng2023pdrf, lee2023exblurf, lee2024deblurring, peng2024bags} have been proposed to reconstruct sharp radiance fields from training views with either camera motion blur or defocus blur.
These radiance field deblurring methods jointly optimize blur kernels and the radiance field, and have shown promising results.

However, these methods still possess several limitations that hinder their effectiveness.
First, existing approaches rely on linear blur models, describing blurred pixels in training views as linear combinations of sharp pixels.
In contrast, real-world blurred images often suffer from non-linear outliers such as saturated pixels and noise, and non-linear in-camera processing~\cite{cho2011handling_original,rim_2022_ECCV_original}, which can severely degrade the performance of current radiance field deblurring techniques.
Second, existing approaches do not utilize priors on sharp images, relying solely on complementary information from different views to model blur kernels and estimate a sharp radiance field. Consequently, they tend to produce radiance fields with residual blur and may completely fail when all input views exhibit similar blur directions~\cite{ma2022deblurnerf}, akin to classical multi-frame deblurring approaches~\cite{chen1996image,rav2005two,cho2007removing}, due to the lack of sharp image priors.
Third, existing ray-based methods~\cite{ma2022deblurnerf, wang2023badnerf, lee2023dpnerf, peng2023pdrf, lee2023exblurf} employ multiple ray samples per pixel to depict a blurry image, resulting in a substantial increase in rendering time for a single view from a radiance field, and consequently, a significant computation time for radiance field construction.

Apart from the radiance field deblurring methods, image deblurring has been extensively studied for decades~\cite{cho2009fast, hirsch2011fast, zhang2013non, pan2014deblurring, whyte10nonuniform}.
Recently, a number of deep neural network (DNN)-based single-image deblurring methods have been proposed~\cite{nah2017gopro, tao2018srn, kupyn2018deblurgan, zamir2021mprnet, cho2021mimounet, chen2022nafnet}.
These methods do not rely on linear blur models, but learn image deblurring from large-scale datasets.
Thanks to this, they can effectively remove blur from a single image without the need for complementary information from other images, and also handle non-linearities such as saturated pixels~\cite{rim2020realblur}.
Moreover, their feed-forward approach allows markedly reduced processing times, unlike classical iterative optimization-based approaches.

Given the effectiveness of recent DNN-based deblurring approaches, a promising direction to construct a sharp radiance field from blurry images would be to combine a deblurring network and radiance field construction.
However, a na\"ive combination of a deblurring network and radiance field construction, which performs single-image deblurring to each input blurry image and trains a radiance field using the deblurred images, results in unsatisfactory results as reported by Ma \etal~\cite{ma2022deblurnerf}.
This is because of the limited performance of single-image deblurring, which stems from the insufficient information available in a single image.

In this paper, we propose \textit{\MethodName{}}, a novel radiance field deblurring framework that enables highly efficient training and high-quality novel-view synthesis from images blurred by motion blur or defocus blur, even in the presence of noise and saturated pixels, as shown in \cref{fig:teaser}. Unlike existing radiance field deblurring approaches, our method leverages DNN-based deblurring modules to achieve superior performance and computational efficiency. Additionally, our approach is versatile, capable of handling various 3D representations such as voxel grids~\cite{fridovich2022plenoxels} and 3D Gaussians~\cite{kerbl2023gaussiansplatting}.

However, solely relying on deblurring networks to remove blur from input blurry images may yield unsatisfactory results as discussed above.
To effectively combine DNN-based deblurring and radiance field construction, our framework adopts a novel radiance field (RF)-guided deblurring scheme, and an iterative framework that performs RF-guided deblurring and radiance field construction in an alternating manner.
Specifically, our framework first deblurs input images, and then trains a radiance field using the deblurred images.
Then, at the next iteration, it renders images corresponding to the input views from the trained radiance field.
While the deblurred images may have limited qualities with inaccurately restored details, the rendered images provide higher-quality details as they are rendered from aggregated information from multiple deblurred images.
Using the rendered images as guidance, our framework then performs RF-guided deblurring to the input blurred images.
Thanks to the guidance, we can obtain higher-quality deblurred images, which we use for radiance field construction again.
Finally, our framework iterates RF-guided deblurring and radiance field construction to gradually enhance the quality of the radiance field.

Training the deblurring modules of our framework requires a large-scale dataset of blurred images paired with ground-truth sharp images for radiance field reconstruction.
However, no such datasets are available as existing radiance field deblurring approaches adopt linear blur model-based approaches, which do not require large-scale training datasets~\cite{ma2022deblurnerf, wang2023badnerf, lee2023dpnerf, lee2023exblurf, peng2023pdrf, lee2024deblurring, peng2024bags}.
Consequently, the datasets provided by previous works, such as those from Deblur-NeRF~\cite{ma2022deblurnerf} and ExBluRF~\cite{lee2023exblurf}, are small and primarily designed for evaluation rather than training. Furthermore, these datasets often overlook crucial factors like camera noise, in-camera processing, and other non-linear outliers, which are essential for realistic deblurring tasks.

To address these gaps, we propose \textit{\SynthDataName{}}, a large-scale dataset for radiance field deblurring approaches. It includes 4,350 blurred-sharp image pairs across 150 scenes, with 2,175 pairs for each of 75 scenes, encompassing both camera motion blur and defocus blur. These images are carefully synthesized to reflect real-world camera degradations such as noise, saturated pixels, and in-camera processing pipelines.
Additionally, we introduce a real-world dataset called \textit{\RealDataName{}} for evaluation in non-ideal conditions. Unlike the Deblur-NeRF dataset, which has less noise and adequate lighting, \textit{\RealDataName{}} consists of five low-light indoor scenes captured with a machine vision camera.

We validate \textit{\MethodName{}} on both synthetic and real-world datasets with two types of blur: camera motion blur and defocus blur.
Experimental results demonstrate that our method achieves state-of-the-art novel-view synthesis performance with highly reduced processing times.
Furthermore, we demonstrate the extensibility of our framework by constructing radiance fields using different scene representations, such as voxel grids and Gaussian Splatting.
Our contributions can be summarized as follows:
\begin{itemize}
    \item We propose \textit{\MethodName{}}, a novel radiance field deblurring framework, which is the first approach that leverages DNN-based deblurring modules to overcome the limitations of the linear blur model.
    \item To this end, we present RF-guided deblurring, and an iterative framework that performs RF-guided deblurring and radiance field construction in an alternating manner to gradually enhance the quality of the radiance field.
    \item We also present the \textit{\SynthDataName{}} dataset, the first large-scale dataset for training and evaluation of novel-view synthesis from blurry images.
\end{itemize}

\section{Related Work}
\label{sec:releated_work}

\paragraph{Radiance field deblurring}
Several works have recently been proposed to tackle radiance field deblurring~\cite{ma2022deblurnerf, wang2023badnerf, lee2023dpnerf, lee2023exblurf, peng2023pdrf}.
Ma \etal~presented Deblur-NeRF, the first framework to train a sharp radiance field from blurry training views~\cite{ma2022deblurnerf}.
Deblur-NeRF adopts an additional MLP to model blur kernels and jointly optimizes them with a radiance field.
DP-NeRF~\cite{lee2023dpnerf} models blur kernels based on physical priors for rigid motions, while PDRF~\cite{peng2023pdrf} uses a two-stage kernel modeling scheme where the first stage coarsely aggregates kernel points to reduce computational cost.
All these methods support both camera motion blur and defocus blur by modeling them using multiple rays.
On the other hand, BAD-NeRF~\cite{wang2023badnerf} and ExBluRF~\cite{lee2023exblurf} propose radiance field deblurring for camera motion blur, modeling camera motion blur using camera trajectories. 
Nevertheless, all these methods rely on linear blur models, and require multiple ray-tracing steps, resulting in limited deblurring quality and slow training speeds.

More recently, with the emergence of 3D Gaussian Splatting~\cite{kerbl2023gaussiansplatting}, 3D Gaussian-based radiance field deblurring methods have been proposed.
Deblurring-3DGS~\cite{lee2024deblurring} handles camera motion blur with small offsets in 3D Gaussian positions and defocus blur by adjusting the geometry of Gaussians. 
BAGS~\cite{peng2024bags} models both types of blur using dense per-pixel blur kernels optimized in a coarse-to-fine manner.
However, BAGS requires substantial computation time to optimize dense blur kernels.
Moreover, these methods also rely on linear blur models and are limited in handling real-world blurred images. 

\paragraph{Image deblurring}
Image deblurring has been widely studied to address blur that degrades image quality. 
Traditional methods, relying on linear blur models, often struggle with real-world blurred images containing non-linear outliers~\cite{cho2009fast, hirsch2011fast, zhang2013non, pan2014deblurring, whyte10nonuniform, cho2012video, cho2007removing, rav2005two}. 
Their iterative optimization processes are also computationally intensive, limiting their practical use. 
Recent advancements in deep learning have introduced DNN-based approaches, including single-image~\cite{nah2017gopro, tao2018srn, kupyn2018deblurgan, zamir2021mprnet, cho2021mimounet, chen2022nafnet} and multi-frame/video deblurring~\cite{wieschollek2017learning, aittala2018burst, zhou2019davanet, gu2020blur, su2017video, nah2019video, deng2021video, hyun2017video, zhou2019video}. 
These methods can handle more complex blurs and reduce processing time with a feed-forward structure.

While applying DNN-based deblurring approaches might be a potential solution for constructing radiance fields from blurry images, na\"ively using single-image deblurring techniques yields unsatisfactory results as discussed in \cref{sec:intro}. 
Another possible direction is to adopt multi-frame and video deblurring approaches that leverage complementary information from multiple images for higher-quality results.
However, these methods assume that input images are captured from nearly the same viewpoints, a condition that does not hold for training views in radiance field construction.

\begin{figure*}
    \centering
    \includegraphics[width=0.98\textwidth]{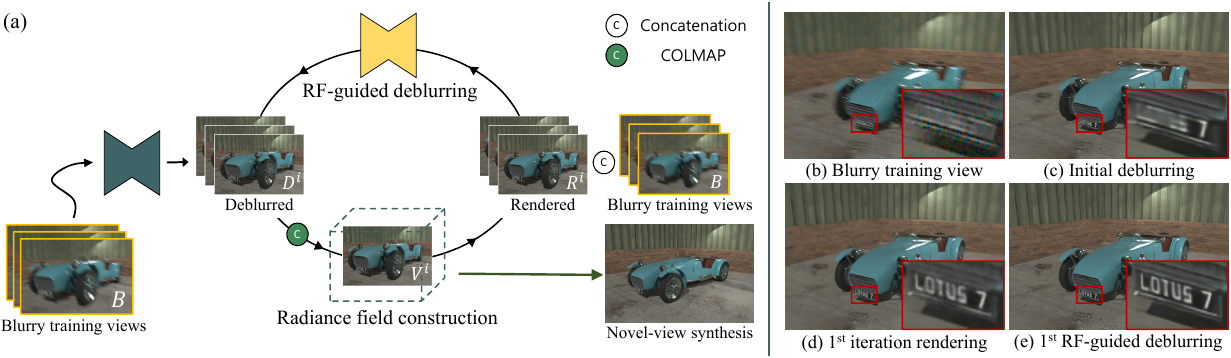}
    \caption{Overall framework and intermediate result of each step of \MethodName{}.}
    \label{fig:overview}
    \vspace{-3mm}
\end{figure*}

\begin{figure*}[t]
    \centering
    \includegraphics[width=0.98\textwidth]{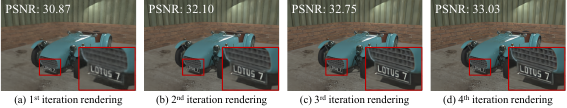}
    \caption{As iterations progress, the rendered images contain increasingly high-quality scene information, which subsequently improves the RF-guided deblurring network's performance in the next iteration.}
    \label{fig:render_iterations}
    \vspace{-3mm}
\end{figure*}

\section{\MethodName{}}
\label{sec:deepdeblurrf}
\cref{fig:overview}-(a) illustrates the overall framework of \MethodName{}.
It takes a set of $M$ blurred images $\textbf{B} = \{B_1, ..., B_M\}$ of a scene and estimates a sharp radiance field that can synthesize a sharp novel view given an arbitrary pose.
We assume that each blurred image $B_m$ is obtained from its latent sharp image $L_m$ through an unknown degradation process including blur and non-linear outliers such as saturated pixels and nonlinear in-camera processing.
Based on this assumption, \MethodName{} first performs initial deblurring to the input blurred images, and obtains initial deblurred images. 
Then, our method iteratively performs radiance field construction using deblurred images and RF-guided deblurring to gradually enhance the quality of the radiance field and the deblurred images.
At the last iteration, we perform only the radiance field construction step and obtain a final radiance field, from which we can synthesize sharp novel views.
In the following, each step of \MethodName{} is described in more detail.

\subsection{Initial Deblurring}
The initial deblurring step removes blur from each blurry training view $B_m \in \textbf{B}$ so that the following radiance field construction step can estimate the pose of each view more accurately, and more effectively aggregate information from different views.
To this end, we adopt an off-the-shelf single-image deblurring network.
Specifically, we adopt NAFNet~\cite{chen2022nafnet}, a state-of-the-art deblurring network, for its computational efficiency and performance.
We denote the deblurred image from the initial deblurring step corresponding to $B_m$ as $D_m^0$, and the set of the deblurred images as $\textbf{D}^0 = \{D_1^0, ..., D_M^0\}$.

\begin{figure*}[t]
    \centering
    \includegraphics[width=0.99\textwidth]{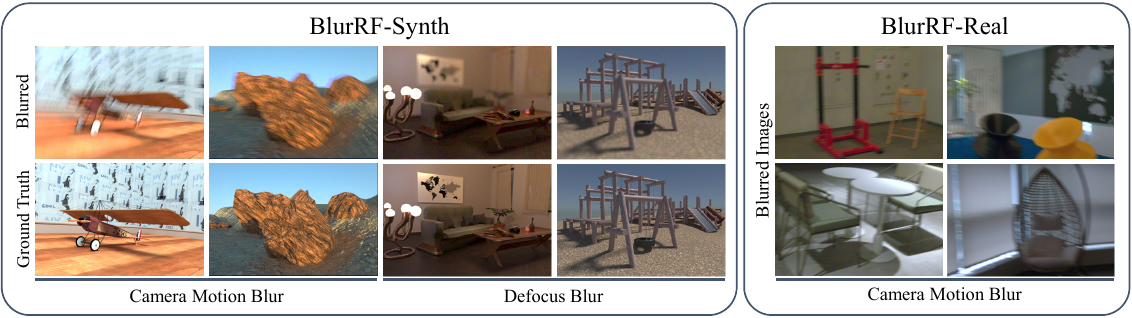}
    \caption{Examples of the \SynthDataName{} and \RealDataName{} datasets. The examples show blurred views in \RealDataName{}, while the top and bottom rows in \SynthDataName{} include blurred views and their corresponding sharp views.}
    \label{fig:dataset}
\end{figure*}

\subsection{Radiance Field Construction}
In the radiance field construction step at the $i$-th iteration, we first estimate the camera poses of the deblurred images from the previous step, denoted as $\textbf{D}^{i-1} = \{D_1^{i-1}, ..., D_M^{i-1}\}$ using COLMAP~\cite{schoenberger2016sfm}, where $i \in [1, N]$ is the index of our iterative process.
Using the estimated poses and deblurred images, we aggregate information about the target 3D scene and construct a radiance field $\mathbf{V}^i$.
As discussed in \cref{sec:intro}, our framework can adopt diverse radiance field representations.
To demonstrate the extensibility of our framework, we adopt two different representations, Plenoxels~\cite{fridovich2022plenoxels} and 3D Gaussians~\cite{kerbl2023gaussiansplatting}, in our experiments.
We refer to our framework using Plenoxels and 3D Gaussians as \MethodName{}-P and \MethodName{}-G, respectively.

Note that deblurred images $\textbf{D}^{i-1}$ from the previous step may still contain residual blur and deblurring artifacts, which can introduce corrupted information into the aggregation process.
Nevertheless, since the input images have overlapping regions, other deblurred images can provide information for those regions that appear corrupted in some images.
By aggregating such information from different deblurred images on the same regions, the radiance field construction step can effectively suppress corrupted information and obtain a higher-quality radiance field, from which we can render higher-quality images compared to the input deblurred images at this step.

\subsection{RF-guided Deblurring}
The RF-guided deblurring step deblurs the input views $\textbf{B}$ using the aggregated information in $\textbf{V}^i$.
Specifically, for each input view $B_m$, we render $\textbf{V}^i$ to obtain a rendered image $R_m^i$.
The rendered image $R_m^i$ shares the same content as its corresponding deblurred image $D_m^{i-1}$, but contains fewer artifacts and finer details.
Motivated by this, our RF-guided deblurring leverages $R_m^i$ to obtain an updated deblurred image $D_m^i$ from $B_m$.

To perform RF-guided deblurring, we employ a novel RF-guided deblurring network that takes both the rendered image $R_m^i$ and the input blurred image $B_m$ to guide the deblurring process with the aggregated information in $R_m^i$.
For the deblurring network, we adopt NAFNet~\cite{chen2022nafnet} and modify its first layer to take a concatenation of the two images.
Once we obtain the updated deblurred images $\textbf{D}^i$, we return to the radiance field construction step and carry out the process for the $(i+1)$-th iteration, gradually enhancing the quality of the radiance field and the deblurred images.

The intermediate results of each step are visualized in \cref{fig:overview}-(b) to \cref{fig:overview}-(e).
Due to the severe blur in the input blurred image, its initial deblurring result retains some residual blur.
In contrast, thanks to the information aggregation occurring in the radiance field construction step, the rendered image contains more accurate details.
Finally, thanks to the guidance of the rendered image, the RF-guided deblurring result shows sharper and more precise details that cannot be observed in the intermediate results of previous steps.

\cref{fig:render_iterations} visualizes rendered images at different iterations.
Although the rendered image at the first iteration \cref{fig:render_iterations}-(a) contains blurry details, they are gradually refined at each iteration (\cref{fig:render_iterations} (b)-(d)) due to our iterative process, which incrementally improves the quality of the radiance fields and the deblurred images.
The effectiveness of our iterative process is also demonstrated by the increasing PSNR score as iterations progress.

\section{BlurRF Dataset}
\label{sec:dataset}

\subsection{\SynthDataName{}}
We propose the \SynthDataName{} dataset, the first large-scale dataset for training and evaluating novel-view synthesis from blurred images.
The dataset comprises a train set and a test set for each type of blur: camera motion blur and defocus blur.
The train set includes 65 scenes, each containing 29 pairs of synthetically blurred images and their corresponding sharp images, captured from different viewpoints. 
The test set includes 10 scenes, each with 29 blurred-sharp image pairs and five additional sharp images from different viewpoints for evaluating novel-view synthesis quality.
Examples of blurred images and ground-truth sharp images are shown in \cref{fig:dataset}.
We synthetically generated the dataset using Blender models, collecting 95 models from Blendswap\footnote{\url{https://blendswap.com/}} under Creative Commons licenses and 5 models from the synthetic dataset of Deblur-NeRF~\cite{ma2022deblurnerf}.
For each model, we sampled 29 camera poses to capture different viewpoints.

\paragraph{Camera motion blur}
We simulated camera shakes by randomly sampling a camera trajectory for each pose during the exposure duration. 
To this end, we adopted B\'ezier interpolation to model 6-DOF camera motion and densely sampled 51 intermediate poses along this camera trajectory.
We then rendered the scene using Blender at each intermediate camera pose to produce a series of sharp images, which we then averaged to create a blurred image.
All rendering and averaging were performed in the linear sRGB color space to accurately replicate the real-world image formation process. 
Among the sharp images, we sampled the temporally central image (i.e., the 26th image) as the ground-truth sharp image.  

\paragraph{Defocus blur}
We also generated images with defocus blur using Blender.
We adjusted the camera’s depth-of-field (DoF) in Blender to create defocus blur.
We controlled the aperture size and randomly set the blade count between 7 and 9 to simulate realistic defocus effects.
For each camera pose, we sampled a random focal distance within a predefined range based on the scene scale, shifting the focal plane to introduce variation in the defocus blur.

\begin{table*}[t]
\centering
\scalebox{0.9}{
        \begin{tabular}{c|l|ccc|ccc|c}
            \Xhline{4\arrayrulewidth}
            \multirow{2}{*}{3D Representation} & \multirow{2}{*}{Model} & \multicolumn{3}{c|}{Camera Motion Blur} & \multicolumn{3}{c|}{Defocus Blur} & Computation \\ 
                                                & & PSNR ($\uparrow$) & SSIM ($\uparrow$) & LPIPS ($\downarrow$) & PSNR ($\uparrow$) & SSIM ($\uparrow$) & LPIPS ($\downarrow$) & Time (Hr.) \\ \hline\hline 
             & Deblur-NeRF~\cite{ma2022deblurnerf} & 27.67 & 0.8340 & 0.1450 & 30.03 & 0.8727 & 0.1137 & 31.33      \\ 
            MLP & BAD-NeRF~\cite{wang2023badnerf}     & 21.74 & 0.5298 & 0.3969 & -     & -      & -      & 24.68      \\ 
             & DP-NeRF~\cite{lee2023dpnerf}        & 28.03 & 0.8412 & 0.1267 & 30.15 & 0.8763 & 0.0991 & 30.00*     \\ \hline
             & ExBluRF~\cite{lee2023exblurf}       & 26.56 & 0.7823 & 0.1955 & -     & -      & -      & 8.92       \\
            Voxel grid & PDRF-10~\cite{peng2023pdrf}         & 28.33 & 0.8435 & 0.1495 & 30.03 & 0.8750 & 0.1225 & 4.26       \\ 
            \rowcolor{skyblue}
             \cellcolor{white} & \cellcolor{white}\textbf{\MethodName{}-P}    & 29.81 & 0.8668 & 0.1142 & 32.51 & 0.9058 & 0.0961 & 1.14\cellcolor{white}              \\ \hline
             & Deblurring-3DGS~\cite{lee2024deblurring}   & 26.30 & 0.7729 & 0.1728 & 29.37 & 0.8545 & 0.1470 & 0.33\cellcolor{skyblue}       \\ 
            3D Gaussians & BAGS~\cite{peng2024bags}            & 27.41 & 0.8108 & 0.1382 & 29.90 & 0.8638 & 0.1152 & 1.25       \\  
            \rowcolor{red_blurry}
            \cellcolor{white} & \cellcolor{white}\textbf{\MethodName{}-G}    & 29.94 & 0.8681 & 0.1059 & 32.58 & 0.9060 & 0.0774 & 0.28       \\ \Xhline{4\arrayrulewidth}
        \end{tabular}
    }
    \caption{Quantitative results of novel-view synthesis on \SynthDataName{} test scenes. We highlight \colorbox{red_blurry}{the best metrics} and \colorbox{skyblue}{the second best metrics}. Note that the computation time of DP-NeRF~\cite{lee2023dpnerf} was measured with two GPUs due to its memory demands, whereas those of the other models were measured on a single GPU.}
    \label{table:quantitative_comparison}
\end{table*}

\begin{figure*}[t]
    \centering
    \includegraphics[width=\textwidth]{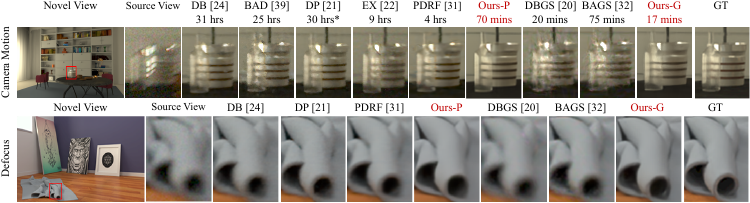}
    \caption{Qualitative results of novel-view synthesis on \SynthDataName{} test scenes.
    }
    \label{fig:qualitative_result1}
\end{figure*}

\paragraph{Realistic blur synthesis}
To reflect real-world image degradation, we adopted the blur synthesis pipeline of RSBlur~\cite{rim_2022_ECCV_original}. 
Specifically, we generated camera motion blur by averaging sharp images in the linear sRGB space.
Then, we synthesized saturated pixels to the blurred images and converted them to the camera RAW space.
In the camera RAW space, we synthesized shot and read noise, and converted the images back to the camera sRGB space.
For defocus blur, we rendered blurred images in the linear sRGB space and added noise in the same manner.
For a more detailed pipeline, we refer the readers to \cite{rim_2022_ECCV_original}.
The blur synthesis pipeline requires camera-specific noise parameters and a color correction matrix.
We used a Sony A7R3 camera to estimate them.
For more details about the \SynthDataName{} dataset, we refer the readers to the supplementary material.

\subsection{\RealDataName{}}
\label{subsec:BlurRF-real}
For evaluation under more realistic and challenging conditions, we propose \RealDataName{}, a real-world low-light camera motion blur dataset. The existing real datasets~\cite{ma2022deblurnerf,lee2023exblurf}  are captured in well-lit environments with minimal noise, making them less representative of camera motion blur in low-light conditions. Our dataset addresses this gap by providing data captured under challenging low-light scenarios. 
We collected five indoor scenes using a machine vision camera, each providing 20-40 multi-view blurry images, including 3-5 images for novel-view synthesis evaluation.
Examples of blurred training views are shown in \cref{fig:dataset}.

\begin{figure*}[t]
    \centering
    \includegraphics[width=\textwidth]{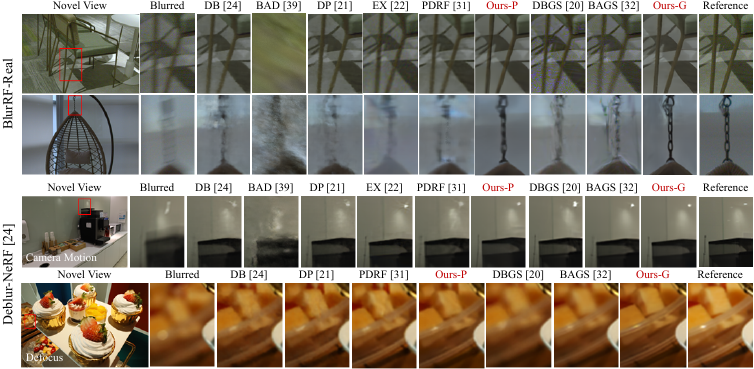}
    \caption{Qualitative results of novel-view synthesis on real-world datasets.
    }
    \label{fig:qualitative_result2}
    \vspace{-2mm}
\end{figure*}

\begin{figure*}[t]
    \centering
    \includegraphics[width=\textwidth]{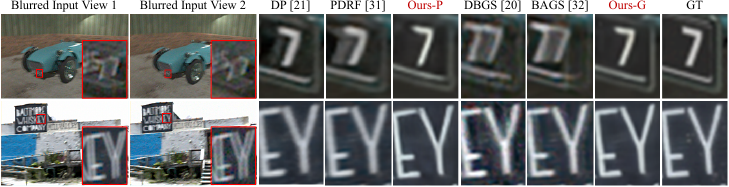}
    \caption{Qualitative results of novel-view synthesis on \SBDataName{} dataset.
    }
    \label{fig:same_blur}
    \vspace{-2mm}
\end{figure*}
\section{Experiments}
\label{sec:experiments}
\paragraph{Datasets}
The deblurring networks of \MethodName{} are trained with the train set of \SynthDataName{}.
For evaluating \MethodName{} and other radiance field deblurring methods, we use the test set of \SynthDataName{}, the real-world dataset from Deblur-NeRF~\cite{ma2022deblurnerf}, and the \RealDataName{} dataset.
The test set of \SynthDataName{} and the real-world dataset from Deblur-NeRF each contain 10 scenes per blur type, while the \RealDataName{} dataset has 5 scenes with camera motion blur.
The real scenes of Deblur-NeRF were captured with a Canon EOS RP under the manual exposure mode.
The camera poses are estimated using COLMAP~\cite{schoenberger2016sfm}.

\paragraph{Implementation details}
\label{Implementation_Details}
We set the number of iterations $N$ to 5.
Thus, as detailed in \cref{sec:deepdeblurrf}, \MethodName{} includes one single-image deblurring network and four RF-guided deblurring networks for each blur type: camera motion blur and defocus blur.
Note that our approach requires an additional training phase for the deblurring networks, unlike previous radiance field deblurring methods.
Once trained, our framework can construct radiance fields for arbitrary scenes and render novel views, similar to conventional approaches.
To train the deblurring networks, we use the Adam optimizer~\cite{kingma2014adam} with $\beta_1 = 0.9$ and $\beta_2 = 0.9$. 
The learning rate is initially set to $10^{-3}$ and gradually reduced to $10^{-7}$ using the cosine annealing scheduler~\cite{loshchilov2016sgdr}.

For constructing radiance fields in \MethodName{}-P, we follow the forward-facing experimental setup in Plenoxels, applying pruning and upsampling to the voxel grids every 38,400 iterations. 
For \MethodName{}-G, we adopt the original Gaussian Splatting method, incorporating the depth-based pruning and additional points strategy from Deblurring-3DGS~\cite{lee2024deblurring} to address the sparse point clouds obtained from COLMAP on blurred images.

We measured the computation time for radiance field deblurring using the camera motion blur test set of \SynthDataName{} on a single NVIDIA TITAN RTX, except for DP-NeRF, which requires two GPUs due to its memory demands.
Additional implementation details, including network configurations and the training settings for each iteration of \MethodName{}-P and \MethodName{}-G, are provided in the supplementary material.

\paragraph{Comparison using BlurRF-Synth}
We compare \MethodName{} with other radiance field deblurring methods ~\cite{ma2022deblurnerf, wang2023badnerf, lee2023dpnerf, lee2023exblurf, peng2023pdrf, lee2024deblurring, peng2024bags}.
\cref{table:quantitative_comparison} shows the quantitative results of \MethodName{} and the other methods on the test set of \SynthDataName{}. 
The table demonstrates that our approach outperforms the existing methods in terms of quality, regardless of the radiance field representation. 
Moreover, it achieves substantially shorter computation times. 
Specifically, \MethodName{}-P shows the shortest computation time among the approaches based on MLPs and voxel grids, while \MethodName{}-G is the fastest among all the methods. 
This superior performance can be attributed to its DNN-based deblurring modules, which handle blur efficiently and effectively, in contrast to the other methods constrained by linear blur models.

The superior performance of our approach is further demonstrated in \cref{fig:qualitative_result1}, which shows that our method achieves sharper results on the \SynthDataName{} dataset.
Specifically, on the camera motion blur scene, all the other methods produce noisy results, whereas both \MethodName{}-P and \MethodName{}-G effectively remove the blur even in challenging conditions with noise and pixel saturation, achieving high-quality novel-view synthesis.
Similarly, in the defocus blur scene, our methods render sharper novel views by minimizing residual blur than the other methods.

\paragraph{Comparison on real datasets}
We compare the novel-view synthesis performance of \MethodName{} against other approaches using both \RealDataName{} and the real scenes from Deblur-NeRF~\cite{ma2022deblurnerf}.
As detailed in \cref{subsec:BlurRF-real}, \RealDataName{} provides an evaluation of how well each method handles camera motion blur with non-linear outliers and noise in real-world low-light conditions.
As shown in \cref{table:BlurRF-SB} and \cref{fig:qualitative_result2}, our method not only outperforms the others in non-reference metrics~\cite{mittal2012completelyblind, agnolucci2024arniqa} but also achieves high-quality qualitative results, clearly surpassing the competing methods.
Even in Deblur-NeRF's real-world scenes, which have minimal noise, our method demonstrates superior performance.
Specifically, while other methods struggle with saturated pixels in the camera motion blur scene, our method effectively handles it.
Likewise, in the defocus blur scene, our method renders sharper novel views without residual blur.

While the real-world scenes of Deblur-NeRF provide blur-free reference images, they suffer from misalignment and exposure differences~\cite{ma2022deblurnerf, peng2023pdrf, peng2024bags}.
Similarly, \RealDataName{} also includes blur-free reference images, but they have not only similar limitations but also severe noise as they are captured under severe low-light conditions.
Thus, we report quantitative evaluations against the reference images in the supplementary material.
More experimental results are also provided in the supplementary material.

\begin{table}[t]
\centering
\scalebox{0.65}{
    \begin{tabular}{l|cc|ccc}
        \Xhline{4\arrayrulewidth}
        \multirow{2}{*}{Model} & \multicolumn{2}{c}{\RealDataName{}} & \multicolumn{3}{|c}{\SBDataName{}} \\
                               & NIQE ($\downarrow$) & ARNIQA ($\uparrow$) & PSNR ($\uparrow$) & SSIM ($\uparrow$) & LPIPS ($\downarrow$) \\ \hline\hline
        \multicolumn{6}{c}{MLP} \\ \hline 
        Deblur-NeRF~\cite{ma2022deblurnerf} & 6.341 & 0.279 & 25.43 & 0.7264 & 0.2319 \\
        BAD-NeRF~\cite{wang2023badnerf}     & 9.908 & 0.237 & 21.97 & 0.5231 & 0.4731 \\
        DP-NeRF~\cite{lee2023dpnerf}        & 6.498 & 0.266 & 27.07 & 0.7940 & 0.1795 \\ \hline
        \multicolumn{6}{c}{Voxel grid} \\ \hline 
        ExBluRF~\cite{lee2023exblurf}       & 7.479 & 0.279 & 24.16 & 0.6649 & 0.2659 \\
        PDRF-10~\cite{peng2023pdrf}         & 6.243 & 0.266 & 26.66 & 0.7730 & 0.2124 \\
        \rowcolor{skyblue}
        \cellcolor{white}\textbf{\MethodName{}-P}                              & 5.829 & 0.323 & 29.45 & 0.8472 & 0.1677 \\ \hline
        \multicolumn{6}{c}{3D Gaussians} \\ \hline 
        Deblurring-3DGS~\cite{lee2024deblurring}  & 6.086 & 0.262 & 23.87 & 0.6459 & 0.2475 \\
        BAGS~\cite{peng2024bags}            & 5.837 & 0.290 & 24.49 & 0.6748 & 0.2391 \\
        \rowcolor{red_blurry}
        \cellcolor{white}\textbf{\MethodName{}-G}                              & 5.423 & 0.329 & 29.59 & 0.8548 & 0.1399 \\ \Xhline{4\arrayrulewidth}
    \end{tabular}
    }
    \caption{Quantitative results of novel-view synthesis on \RealDataName{} and \SBDataName{} datasets. We highlight \colorbox{red_blurry}{the best metrics} and \colorbox{skyblue}{the second best metrics}.}
    \vspace{-3mm}
    \label{table:BlurRF-SB}
\end{table}

\paragraph{Input views with same blur directions}
As discussed in \cref{sec:intro}, unlike previous approaches, our method does not rely on the assumption that input blurred images contain complementary information from different blur directions.
To verify this, we generated an additional synthetic test set, named \SBDataName{}, where the blurred images of the same scene share the same blur directions, while their blur magnitudes are different.
\SBDataName{} was generated using five scenes from the \SynthDataName{} test set, following the same generation process.

\cref{fig:same_blur} shows a qualitative comparison on \SBDataName{}.
Due to the lack of complementary information in different training views, the other methods fail to restore sharp details.
Conversely, our methods successfully restore sharp details.
The quantitative comparison in \cref{table:BlurRF-SB} also demonstrates that our approach significantly outperforms the others for blurred images with the same blur directions.
These results verify the benefit of using prior knowledge on sharp images from pre-trained deblurring networks.

\begin{table}[t]
\centering
\scalebox{0.85}{
        \begin{tabular}{c|ccc|ccc}
            \Xhline{4\arrayrulewidth}
            \multirow{2}{*}{\# Iter.} & \multicolumn{3}{c|}{Camera motion} & \multicolumn{3}{c}{Defocus}  \\
                    & PSNR  & SSIM   & LPIPS  & PSNR  & SSIM   & LPIPS \\ \hline\hline
            $N = 1$ & 28.40 & 0.8175 & 0.1707 & 30.74 & 0.8862 & 0.1297 \\ 
            $N = 2$ & 29.23 & 0.8462 & 0.1281 & 31.22 & 0.8905 & 0.0955 \\ 
            $N = 3$ & 29.65 & 0.8588 & 0.1152 & 32.12 & 0.9003 & 0.0899 \\ 
            $N = 4$ & 29.83 & 0.8632 & 0.1108 & 32.30 & 0.9038 & 0.0860 \\
            $N = 5$ & 29.94 & 0.8681 & 0.1059 & 32.58 & 0.9060 & 0.0774 \\ 
            $N = 6$ & 29.97 & 0.8692 & 0.1034 & 32.75 & 0.9087 & 0.0736  \\ \Xhline{4\arrayrulewidth}
        \end{tabular}
    }
    \caption{Quantitative results of ablation study on the number of iterations $N$ using \MethodName{}-G on the test sets of \SynthDataName{}.}
    \label{table:ablation_cycle}
    \vspace{-3mm}
\end{table}

\begin{figure}[t]
    \centering
    \includegraphics[width=0.48\textwidth]{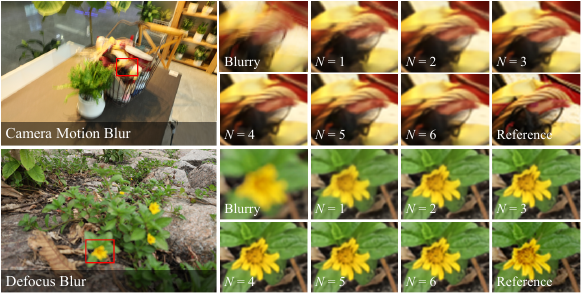}
    \caption{Qualitative results of ablation study on the number of iterations $N$ using \MethodName{}-G on the real-world scenes~\cite{ma2022deblurnerf}.
    }
    \vspace{-3mm}
    \label{fig:ablation_cycle}
\end{figure}

\paragraph{Number of iterations}
We investigate the impact of the hyperparameter $N$, which sets the number of iterations in our framework.
\cref{table:ablation_cycle} shows that \MethodName{}-G achieves improved novel-view synthesis as $N$ increases, highlighting the effectiveness of our iterative approach.
However, gains diminish beyond $N = 5$, suggesting that $N = 5$ is sufficient.
\cref{fig:ablation_cycle} further supports this with qualitative results on real scenes~\cite{ma2022deblurnerf} (camera motion blur and defocus blur), showing similar trends.
Similar results were observed for \MethodName{}-P, as detailed in the supplementary material.

\section{Conclusion}
\label{sec:conclusion}
In this paper, we proposed \MethodName{}, a novel radiance field deblurring approach that leverages DNN-based deblurring to overcome the limitations of the linear blur model of previous methods.
To integrate DNN-based deblurring with radiance field construction, we presented a novel RF-guided deblurring scheme and the framework that alternates between RF-guided deblurring and radiance field training.
We also presented \SynthDataName{}, the first large-scale dataset for training and evaluating radiance field deblurring frameworks.
Experimental results showed that our method outperforms others, handling both camera motion blur and defocus blur effectively with highly reduced computation time.

\paragraph{Limitations and future work}
Our method leverages deblurring networks to remove blur from training views, but restoring images with more severe blur than in the training data remains challenging. 
Our RF-guided deblurring uses only rendered images from each input view for guidance. 
Imposing additional priors on other views and incorporating additional information, e.g., depth maps rendered from radiance fields, may further improve the deblurring performance.

\paragraph{Acknowledgements}
This work was supported by Institute of Information \& communications Technology Planning \& Evaluation (IITP) grants funded by the Korea government(MSIT) (RS-2024-00457882, AI Research Hub Project, RS-2019-II191906, Artificial Intelligence Graduate School Program(POSTECH), RS-2024-00437866, ITRC(Information Technology Research Center)).
{
    \small
    \bibliographystyle{ieeenat_fullname}
    \bibliography{main}
}


\end{document}